\def\year{2022}\relax
\documentclass[letterpaper]{article} 
\usepackage{aaai22}  
\usepackage{times}  
\usepackage{helvet}  
\usepackage{courier}  
\usepackage[hyphens]{url}  
\usepackage{graphicx} 
\usepackage{natbib}  
\usepackage{caption} 
\DeclareCaptionStyle{ruled}{labelfont=normalfont,labelsep=colon,strut=off} 
\usepackage{algorithm}
\usepackage{algorithmic}

\usepackage{newfloat}
\usepackage{listings}
\floatstyle{ruled}
\newfloat{listing}{tb}{lst}{}
\floatname{listing}{Listing}

\usepackage{float}
\usepackage{multirow}
\usepackage{amsmath} 
\usepackage{amssymb}
\usepackage{subcaption}

\title{TransFG: A Transformer Architecture for Fine-Grained Recognition}
\author{
	Ju He$^{1}$
	\;\; Jie-Neng Chen$^1$
	\;\; Shuai Liu$^2$\\
	\;\; Adam Kortylewski$^1$
	\;\; Cheng Yang$^2$
	\;\; Yutong Bai$^1$
	\;\; Changhu Wang$^2$\\
	$^1$Johns Hopkins University \;\; $^2$ByteDance Inc. \\	
}

\begin{document}

\maketitle

\begin{abstract}
    Fine-grained visual classification (FGVC) which aims at recognizing objects from subcategories is a very challenging task due to the inherently subtle inter-class differences. Most existing works mainly tackle this problem by reusing the backbone network to extract features of detected discriminative regions. However, this strategy inevitably complicates the pipeline and pushes the proposed regions to contain most parts of the objects thus fails to locate the really important parts. Recently, vision transformer (ViT) shows its strong performance in the traditional classification task. The self-attention mechanism of the transformer links every patch token to the classification token. In this work, we first evaluate the effectiveness of the ViT framework in the fine-grained recognition setting. Then motivated by the strength of the attention link can be intuitively considered as an indicator of the importance of tokens, we further propose a novel Part Selection Module that can be applied to most of the transformer architectures where we integrate all raw attention weights of the transformer into an attention map for guiding the network to effectively and accurately select discriminative image patches and compute their relations. A contrastive loss is applied to enlarge the distance between feature representations of confusing classes. We name the augmented transformer-based model TransFG and demonstrate the value of it by conducting experiments on five popular fine-grained benchmarks where we achieve state-of-the-art performance. Qualitative results are presented for better understanding of our model. 
\end{abstract}
\section{Introduction}

\begin{figure}
    \centering
    \includegraphics[width=\linewidth,height=5cm]{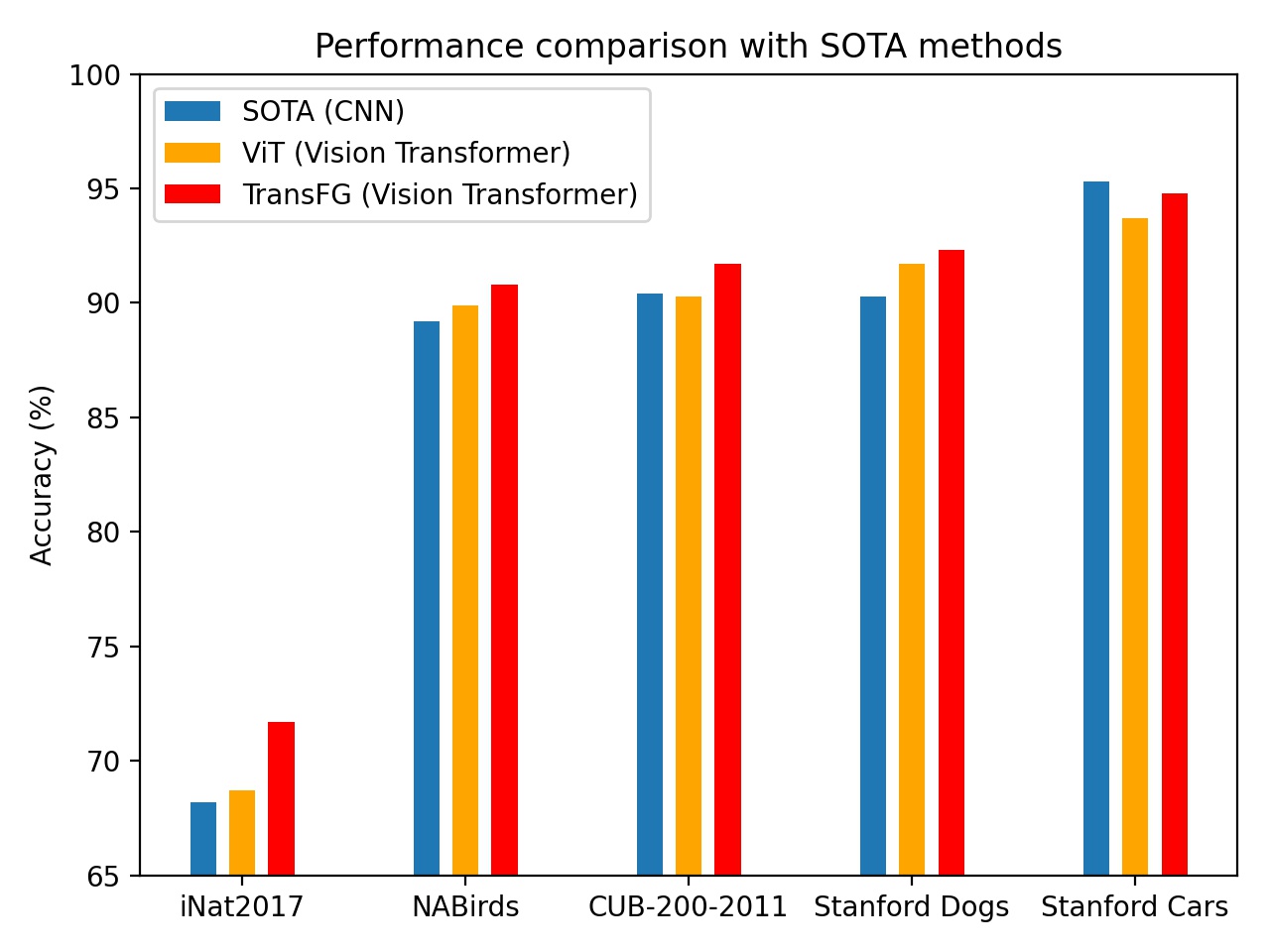}
    \caption{An overview of performance comparison of ViT and TransFG with state-of-the-art methods CNNs on five datasets. We achieve state-of-the-art performance on most datasets while performing a little bit worse on Stanford Cars possibly due to the more regular and simpler car shapes.}
    \label{fig:intro}
\end{figure}

Fine-grained visual classification aims at classifying sub-classes of a given object category, e.g., subcategories of birds \cite{WahCUB_200_2011, van2015building}, cars \cite{KrauseStarkDengFei-Fei_3DRR2013}, aircrafts \cite{maji2013fine}. It has long been considered as a very challenging task due to the small inter-class variations and large intra-class variations along with the deficiency of annotated data, especially for the long-tailed classes. Benefiting from the progress of deep neural networks \cite{NIPS2012_c399862d, simonyan2014very, he2016deep}, the performance of FGVC has obtained a steady progress in recent years. To avoid labor-intensive part annotation, the community currently focuses on weakly-supervised FGVC with only image-level labels. Methods now can be roughly classified into two categories, i.e., localization methods and feature-encoding methods. Compared to feature-encoding methods, the localization methods have the advantage that they explicitly capture the subtle differences among sub-classes which is more interpretable and yields better results.

Early works in localization methods rely on the annotations of parts to locate discriminative regions while recent works \cite{ge2019weakly, liu2020filtration, ding2019selective} mainly adopt region proposal networks (RPN) to propose bounding boxes which contain the discriminative regions. After obtaining the selected image regions, they are resized into a predefined size and forwarded through the backbone network again to acquire informative local features. A typical strategy is to use these local features for classification individually and adopt a rank loss \cite{chen2009ranking} to maintain consistency between the quality of bounding boxes and their final probability output. However, this mechanism ignores the relation between selected regions and thus inevitably encourages the RPN to propose large bounding boxes that contain most parts of the objects which fails to locate the really important regions. Sometimes these bounding boxes can even contain large areas of background and lead to confusion. Additionally, the RPN module with different optimizing goals compared to the backbone network makes the network harder to train and the re-use of backbone complicates the overall pipeline.

Recently, the vision transformer \cite{dosovitskiy2020image} achieved huge success in the classification task which shows that applying a pure transformer directly to a sequence of image patches with its innate attention mechanism can capture the important regions in images. A series of extended works on downstream tasks such as object detection \cite{carion2020end} and semantic segmentation \cite{zheng2021rethinking, xie2021trans2seg, chen2021transunet} confirmed the strong ability for it to capture both global and local features.

These abilities of the Transformer make it innately suitable for the FGVC task as the early long-range ``receptive field" \cite{dosovitskiy2020image} of the Transformer enables it to locate subtle differences and their spatial relation in the earlier processing layers. In contrast, CNNs mainly exploit the locality property of image and only capture weak long-range relation in very high layers. Besides, the subtle differences between fine-grained classes only exist in certain places thus it is unreasonable to convolve a filter which captures the subtle differences to all places of the image.

Motivated by this opinion, in the paper, we present the first study which explores the potential of vision transformers in the context of fine-grained visual classification. We find that directly applying ViT on FGVC already produces satisfactory results while a lot of adaptations according to the characteristics of FGVC can be applied to further boost the performance. To be specific, we propose Part Selection Module which can find the discriminative regions and remove redundant information. A contrastive loss is introduced to make the model more discriminative. We name this novel yet simple transformer-based framework TransFG, and evaluate it extensively on five popular fine-grained visual classification benchmarks (CUB-200-2011, Stanford Cars, Stanford Dogs, NABirds, iNat2017). An overview of the performance comparison can be seen in Fig \ref{fig:intro} where our TransFG outperforms existing SOTA CNN methods with different backbones on most datasets. In summary, we make several important contributions in this work:

\begin{enumerate}
    \item To the best of our knowledge, we are the first to verify the effectiveness of vision transformer on fine-grained visual classification which offers an alternative to the dominating CNN backbone with RPN model design.
    \item We introduce TransFG, a novel neural architecture for fine-grained visual classification that naturally focuses on the most discriminative regions of the objects and achieve SOTA performance on several benchmarks.
    \item Visualization results are presented which illustrate the ability of our TransFG to accurately capture discriminative image regions and help us to better understand how it makes correct predictions.
\end{enumerate}
\begin{figure*}
    \centering
    \includegraphics[width=\linewidth,height=8cm]{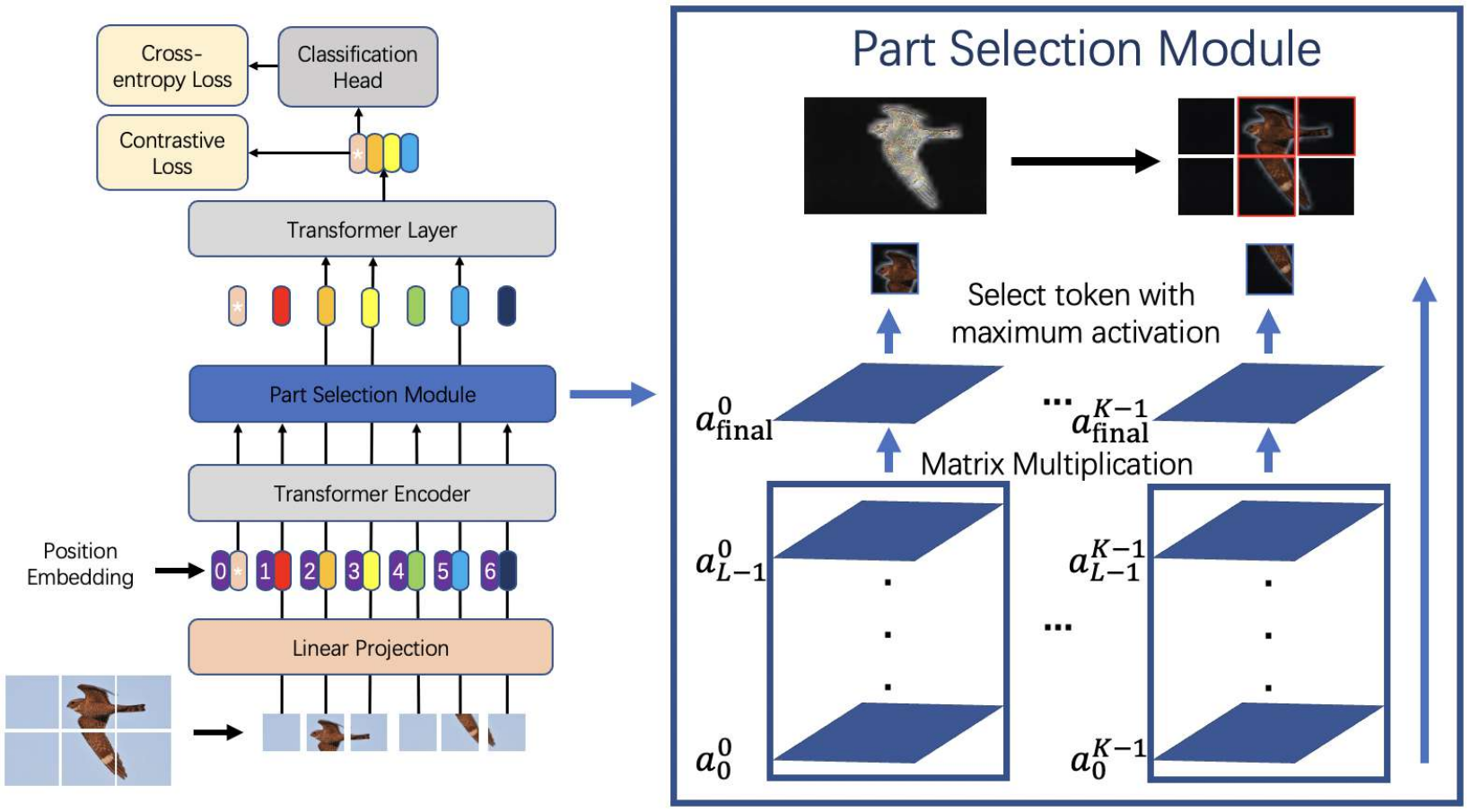}
    \caption{The framework of our proposed TransFG. Images are split into small patches (a non-overlapping split is shown here) and projected into the embedding space. The input to the Transformer Encoder consists of patch embeddings along with learnable position embeddings. Before the last Transformer Layer, a Part Selection Module (PSM) is applied to select tokens that corresponds to the discriminative image patches and only use these selected tokens as input. Best viewed in color.}
    \label{fig:model}
\end{figure*}

\section{Related Work}

In this section, we briefly review existing works on fine-grained visual classification and transformer.

\subsection{Fine-Grained Visual Classification}
Many works have been done to tackle the problem of fine-grained visual classification and they can roughly be classified into two categories: localization methods \cite{ge2019weakly, liu2020filtration, yang2021re} and feature-encoding methods \cite{yu2018hierarchical, zheng2019learning, gao2020channel}. The former focuses on training a detection network to localize discriminative part regions and reuse them to perform classification. The latter targets at learning more informative features by either computing higher-order information or finding the relationships among contrastive pairs.

\subsubsection{Localization FGVC methods}
Previously, some works \cite{branson2014bird, wei2016mask} tried to exploit the part annotations to supervise the learning procedure of the localization process. However, since such annotations are expensive and usually unavailable, weakly-supervised parts proposal with only image-level labels draw more attentions nowadays. Ge et al. \cite{ge2019weakly} exploited Mask R-CNN and CRF-based segmentation alternatively to extract object instances and discriminative regions. Yang et al. \cite{yang2021re} proposed a re-ranking strategy to re-rank the global classification results based on the database constructed with region features. However, these methods all need a special designed module to propose potential regions and these selected regions need to be forwarded through the backbone again for final classification which is not required in our model and thus keeps the simplicity of our pipeline. 

\subsubsection{Feature-encoding methods}
The other branch of methods focus on enriching the feature representations to obtain better classification results. Yu et al. \cite{yu2018hierarchical} proposed a hierarchical framework to do cross-layer bilinear pooling. Zheng et al. \cite{zheng2019learning} adopted the idea of group convolution to first split channels into different groups by their semantic meanings and then do the bilinear pooling within each group without changing the dimension thus it can be integrated into any existed backbones directly. However, these methods are usually not interpretable such one does not know what makes the model distinguish sub-categories with subtle differences while our model drops unimportant image patches and only keeps those that contain most information for the fine-grained recognition.

\subsection{Transformer} 
Transformer and self-attention models have greatly facilitated research in natural language processing and machine translation \cite{dai2019transformer, devlin2018bert, vaswani2017attention}. Inspired by this, many recent studies try to apply transformers in computer vision area. Initially, transformer is used to handle sequential features extracted by CNN backbone for the videos \cite{girdhar2019video}. Later, transformer models are further extended to other popular computer vision tasks such as object detection \cite{carion2020end, zhu2020deformable}, segmentation \cite{xie2021trans2seg, wang2021max}, object tracking \cite{sun2020transtrack}.
Most recently, pure transformer models are becoming more and more popular. ViT \cite{dosovitskiy2020image} is the first work to show that applying a pure transformer directly to a sequence of image patches can yield state-of-the-art performance on image classification. Based on that, Zheng et al. \cite{zheng2021rethinking} proposed SETR to exploit ViT as the encoder for segmentation. He et al. \cite{he2021transreid} proposed TransReID which embedded side information into transformer along with the JPM to boost the performance on object re-identification. In this work, we extend ViT to fine-grained visual classification and show its effectiveness. 
\section{Method}

We first briefly review the framework of vision transformer and show how to do some preprocessing steps to extend it into fine-grained recognition. Then, the overall framework of TransFG will be elaborated.

\subsection{Vision transformer as feature extractor}
\label{sec:ViT}

\textbf{Image Sequentialization.} Following ViT, we first preprocess the input image into a sequence of flattened patches $x_p$. However, the original split method cut the images into non-overlapping patches, which harms the local neighboring structures especially when discriminative regions are split. To alleviate this problem, we propose to generate overlapping patches with sliding window. To be specific, we denote the input image with resolution $H * W$, the size of image patch as $P$ and the step size of sliding window as $S$. Thus the input images will be split into N patches where
\begin{align}
\label{equ:split}
    N = N_H * N_W = \lfloor \frac{H - P + S}{S} \rfloor * \lfloor \frac{W - P + S}{S} \rfloor
\end{align}
In this way, two adjacent patches share an overlapping area of size $(P - S) * P$ which helps to preserve better local region information. Typically speaking, the smaller the step $S$ is, the better the performance will be. But decreasing S will at the same time requires more computational cost, so a trade-off needs to be made here.

\noindent \textbf{Patch Embedding.} We map the vectorized patches $x_p$ into a latent D-dimensional embedding space using a trainable linear projection. A learnable position embedding is added to the patch embeddings to retain positional information as follows:
\begin{align}
    \mathbf{z}_0 = [x_p^1\mathbf{E},x_p^2\mathbf{E},\cdots,x_p^N\mathbf{E}] + \mathbf{E}_{pos}
\end{align}
where $N$ is the number of image patches, $\mathbf{E} \in \mathbb{R}^{(P^2 \cdot C) * D}$ is the patch embedding projection, and $\mathbf{E}_{pos} \in \mathbb{R}^{N * D}$ denotes the position embedding.

The Transformer encoder \cite{vaswani2017attention} contains $L$ layers of multi-head self-attention (MSA) and multi-layer perceptron (MLP) blocks. Thus the output of the $l$-th layer can be written as follows:
\begin{flalign}
    & \mathbf{z}^{'}_{l} = MSA(LN(\mathbf{z}_{l-1})) + \mathbf{z}_{l-1} & l \in 1, 2, \cdots, L \\ 
    & \mathbf{z}_{l} = MLP(LN(\mathbf{z}^{'}_{l})) + \mathbf{z}^{'}_{l} & l \in 1, 2, \cdots, L 
\end{flalign}
where $LN(\cdot)$ denotes the layer normalization operation and $\mathbf{z}_l$ is the encoded image representation. ViT exploits the first token of the last encoder layer $z_{L}^{0}$ as the representation of the global feature and forward it to a classifier head to obtain the final classification results without considering the potential information stored in the rest of the tokens.

\subsection{TransFG Architecture}
\label{sec:TransFG}

While our experiments show that the pure Vision Transformer can be directly applied into fine-grained visual classification and achieve impressive results, it does not well capture the local information required for FGVC. To this end, we propose the Part Selection Module (PSM) and apply contrastive feature learning to enlarge the distance of representations between confusing sub-categories. The framework of our proposed TransFG is illustrated in Fig \ref{fig:model}.

\subsubsection{Part Selection Module}

\begin{figure}
    \centering
    \includegraphics[width=\linewidth,height=6cm]{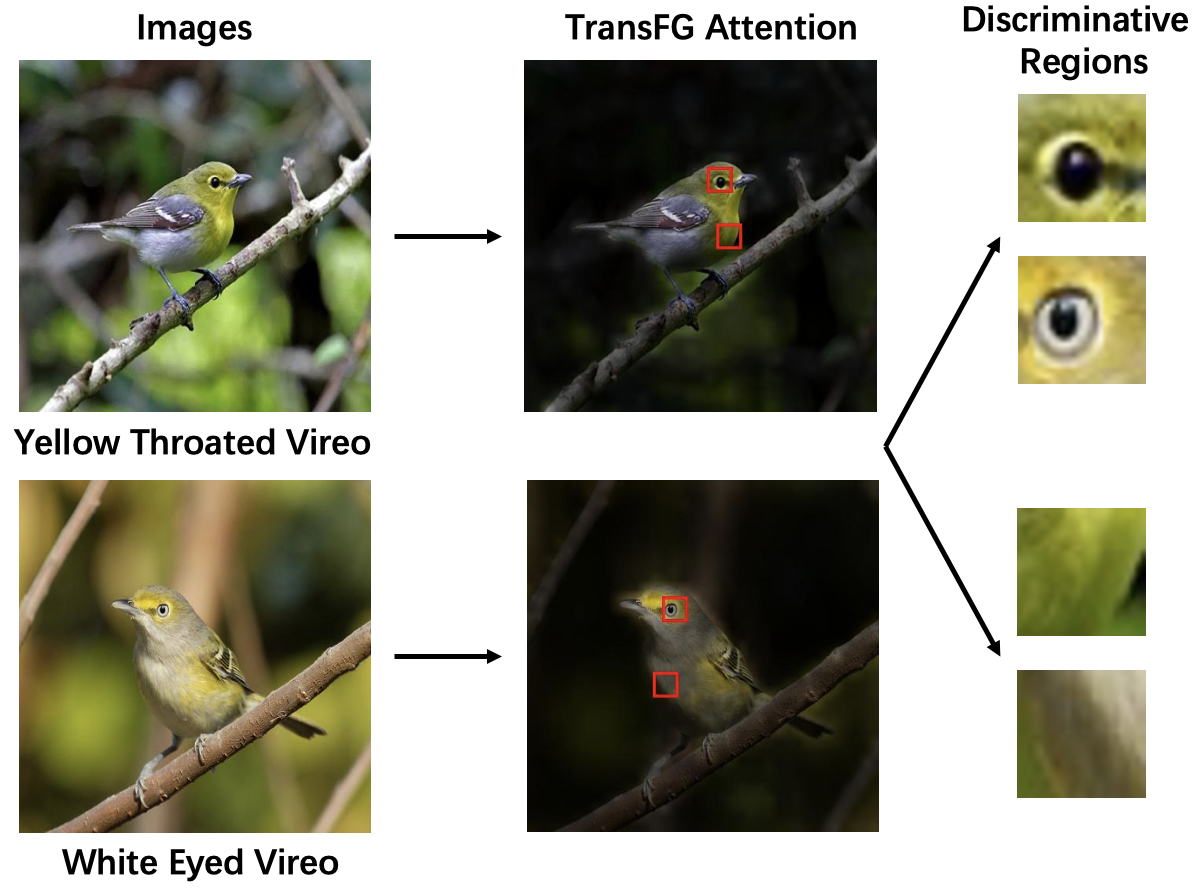}
    \caption{A confusing pair of instances from the CUB-200-2011 dataset. Model needs to has the ability to capture the subtle differences in order to classify them correctly. The second column shows the overall attention maps and two selected tokens of our TransFG method. Best viewed in color.}
    \label{fig:dataset}
\end{figure}

One of the most important problems in fine-grained visual classification is to accurately locate the discriminative regions that account for subtle differences between similar sub-categories. For example, Fig 3 shows a confusing pair of images from the CUB-200-2011 (citation) dataset. The model needs to have the ability to capture the very small differences, i.e., the color of eyes and throat in order to distinguish these two bird species. Region proposal networks and weakly-supervised segmentation strategies are widely introduced to tackle this problem in the traditional CNN-based methods.

Vision Transformer model is perfectly suited here with its innate multi-head attention mechanism. To fully exploit the attention information, we change the input to the last Transformer Layer. Suppose the model has K self-attention heads and the hidden features input to the last layer are denoted as $\mathbf{z}_{L-1}=[z_{L-1}^{0};z_{L-1}^{1},z_{L-1}^{2},\cdots,z_{L-1}^{N}]$. The attention weights of the previous layers can be written as follows:
\begin{flalign}
    & \mathbf{a}_l=[a_{l}^{0},a_{l}^{1},a_{l}^{2},\cdots,a_{l}^{K}] & l \in 1, 2, \cdots, L - 1 \\
    & a_{l}^{i} = [a_{l}^{i_{0}};a_{l}^{i_{1}},a_{l}^{i_{2}},\cdots,a_{l}^{i_{N}}] & i \in 0, 1, \cdots, K - 1
\end{flalign}
Previous works \cite{serrano2019attention, abnar2020quantifying} suggested that the raw attention weights do not necessarily correspond to the relative importance of input tokens especially for higher layers of a model, due to lack of token identifiability of the embeddings. To this end, we propose to integrate attention weights of all previous layers. To be specific, we recursively apply a matrix multiplication to the raw attention weights in all the layers as
\begin{align}
    \mathbf{a}_{final} = \prod_{l=0}^{L-1}\mathbf{a}_l
\end{align}
As $\mathbf{a}_{final}$ captures how information propagates from the input layer to the embeddings in higher layers, it serves as a better choice for selecting discriminative regions compared to the single layer raw attention weights ${a}_{L-1}$. We then choose the index of the maximum value $A_1, A_2, \cdots, A_K$ with respect to the K different attention heads in $\mathbf{a}_{final}$. These positions are used as index for our model to extract the corresponding tokens in $\mathbf{z}_{L-1}$. Finally, we concatenate the selected tokens along with the classification token as the input sequence which is denoted as:
\begin{align}
    \mathbf{z}_{local} = [z_{L-1}^{0};z_{L-1}^{A_{1}},z_{L-1}^{A_{2}},\cdots,z_{L-1}^{A_{K}}]
\end{align}
By replacing the original entire input sequence with tokens corresponding to informative regions and concatenate the classification token as input to the last Transformer Layer, we not only keep the global information but also force the last Transformer Layer to focus on the subtle differences between different sub-categories while abandoning less discriminative regions such as background or common features among a super class.


    





\subsubsection{Contrastive feature learning}
Following ViT, we still adopt the first token $z_i$ of the PSM module for classification. A simple cross-entropy loss is not enough to fully supervise the learning of features since the differences between sub-categories might be very small. To this end, we adopt contrastive loss $\mathcal{L}_{con}$ which minimizes the similarity of classification tokens corresponding to different labels and maximizes the similarity of classification tokens of samples with the same label $y$. To prevent the loss being dominated by easy negatives (different class samples with little similarity), a constant margin $\alpha$ is introduced that only negative pairs with similarity larger than $\alpha$ contribute to the loss $\mathcal{L}_{con}$. Formally, the contrastive loss over a batch of size $B$ is denoted as:
\begin{equation}
\label{equ:con}
    \begin{aligned}
    \mathcal{L}_{con} = \frac{1}{B^2}\sum_i^B[\sum_{j:y_i=y_j}^{B}(1-Sim(z_i,z_j)+ \\ \sum_{j:y_i \neq y_j}^{B}\max((Sim(z_i,z_j)-\alpha), 0)]
\end{aligned}
\end{equation}
where $z_i$ and $z_j$ are pre-processed with $l2$ normalization and $Sim(z_i,z_j)$ is thus the dot product of $z_i$ and $z_j$.

In summary, our model is trained with the sum of cross-entropy loss $L_{cross}$ and contrastive $L_{con}$ together which can be expressed as:
\begin{align}
    \mathcal{L} = \mathcal{L}_{cross}(y, y') + \mathcal{L}_{con}(z)
\end{align}
where $\mathcal{L}_{cross}(y,y')$ is the cross-entropy loss between the predicted label $y'$ and the ground-truth label $y$.

\section{Experiments}
\label{sec:exp}

In this section, we first introduce the detailed setup including datasets and training hyper-parameters. Quantitative analysis is then given followed by ablation studies. We further give qualitative analysis and visualization results to show the interpretability of our model.

\subsection{Experiments Setup}
\label{sec:setup}

\textbf{Datasets.} We evaluate our proposed TransFG on five widely used fine-grained benchmarks, i.e., CUB-200-2011 \cite{WahCUB_200_2011}, Stanford Cars \cite{KrauseStarkDengFei-Fei_3DRR2013}, Stanford Dogs \cite{KhoslaYaoJayadevaprakashFeiFei_FGVC2011}, NABirds \cite{van2015building} and iNat2017 \cite{van2018inaturalist}. We also exploit its usage in large-scale challenging fine-grained competitions.

\noindent \textbf{Implementation details.}
Unless stated otherwise, we implement TransFG as follows. First, we resize input images to $448 * 448$ except $304 * 304$ on iNat2017 for fair comparison (random cropping for training and center cropping for testing). We split image to patches of size 16 and the step size of sliding window is set to be 12. Thus the $H, W, P, S$ in Eq \ref{equ:split} are 448, 448, 16, 12 respectively. The margin $\alpha$ in Eq \ref{equ:con} is set to be 0.4. We load intermediate weights from official ViT-B\_16 model pretrained on ImageNet21k. The batch size is set to 16. SGD optimizer is employed with a momentum of 0.9. The learning rate is initialized as 0.03 except 0.003 for Stanford Dogs dataset and 0.01 for iNat2017 dataset. We adopt cosine annealing as the scheduler of optimizer.

All the experiments are performed with four Nvidia Tesla V100 GPUs using the PyTorch toolbox and APEX.

\subsection{Quantitative Analysis}
\label{sec:quan}

We compare our proposed method TransFG with state-of-the-art works on above mentioned fine-grained datasets. The experiment results on CUB-200-2011 and Stanford Cars are shown in Table \ref{tab:cub}. From the results, we find that our method outperforms all previous methods on CUB dataset and achieve competitive performance on Stanford Cars.

\begin{table}[]
    \small
    \centering
    \caption{Comparison of different methods on CUB-200-2011, Stanford Cars.}
    \label{tab:cub}
    \begin{tabular}{c|c|c|c}
    \hline
    Method & Backbone & CUB & Cars \\ \hline
    ResNet-50 & ResNet-50 & 84.5 & - \\
    NTS-Net & ResNet-50 & 87.5 & 93.9 \\ 
    Cross-X & ResNet-50 & 87.7 & 94.6 \\
    DBTNet & ResNet-101 & 88.1 & 94.5 \\
    FDL & DenseNet-161 & 89.1 & 94.2 \\
    PMG & ResNet-50 & 89.6 & 95.1 \\ 
    API-Net & DenseNet-161 & 90.0 & \textbf{95.3} \\
    StackedLSTM & GoogleNet & 90.4 & - \\ \hline
    DeiT & DeiT-B & 90.0 & 93.9 \\
    ViT & ViT-B\_16 & 90.3 & 93.7 \\ 
    TransFG & ViT-B\_16 & \textbf{91.7} & 94.8 \\ \hline
    \end{tabular}
\end{table}

To be specific, the third column of Table \ref{tab:cub} shows the comparison results on CUB-200-2011. Compared to the best result StackedLSTM \cite{Ge_2019_CVPR} up to now, our TransFG achieves a \textbf{1.3\%} improvement on Top-1 Accuracy metric and 1.4\% improvement compared to our base framework ViT \cite{dosovitskiy2020image}. Multiple ResNet-50 are adopted as multiple branches in \cite{ding2019selective} which greatly increases the complexity. It is also worth noting that StackLSTM is a very messy multi-stage training model which hampers the availability in practical use, while our TransFG maintains the simplicity.

The fourth column of Table \ref{tab:cub} shows the results on Stanford Cars. Our method outperforms most existing methods while performs worse than PMG \cite{du2020fine} and API-Net \cite{zhuang2020learning} with small margin. We argue that the reason might be the much more regular and simpler shape of cars. However, even with this property, our TransFG consistently gets \textbf{1.1\%} improvement compared to the standard ViT model.

\begin{table}[]
    \small
    \centering
    \caption{Comparison of different methods on Stanford Dogs.}
    \label{tab:dog}
    \begin{tabular}{c|c|c}
    \hline
    Method & Backbone & Dogs \\ \hline
    MaxEnt & DenseNet-161 & 83.6 \\ 
    FDL & DenseNet-161 & 84.9 \\
    Cross-X & ResNet-50 & 88.9 \\
    API-Net & ResNet-101 & 90.3 \\ \hline
    ViT & ViT-B\_16 & 91.7 \\ 
    TransFG & ViT-B\_16 & \textbf{92.3} \\ \hline
    \end{tabular}
\end{table}

The results of experiments on Stanford Dogs are shown in Table \ref{tab:dog}. Stanford Dogs is a more challenging dataset compared to Stanford Cars with its the more subtle differences between certain species and the large variances of samples from the same category. Only a few methods have tested on this dataset and our TransFG outperforms all of them. While ViT \cite{dosovitskiy2020image} outperforms other methods by a large margin, our TransFG achieves 92.3\% accuracy which outperforms SOTA by \textbf{2.0\%} with its discriminative part selection and contrastive loss supervision.

\begin{table}[]
    \small
    \centering
    \caption{Comparison of different methods on NABirds.}
    \label{tab:na}
    \begin{tabular}{c|c|c}
    \hline
    Method & Backbone & NABirds \\ \hline
    Cross-X & ResNet-50 & 86.4 \\ 
    API-Net & DenseNet-161 & 88.1 \\ 
    CS-Parts & ResNet-50 & 88.5 \\ 
    FixSENet-154 & SENet-154 & 89.2 \\ \hline
    ViT & ViT-B\_16 & 89.9 \\
    TransFG & ViT-B\_16 & \textbf{90.8} \\ \hline
    \end{tabular}
\end{table}

NABirds is a much larger birds dataset not only from the side of images numbers but also with 355 more categories which significantly makes the fine-grained visual classification task more challenging. We show our results on it in Table \ref{tab:na}. 
We observe that most methods achieve good results by either exploiting multiple backbones for different branches or adopting quite deep CNN structures to extract better features. 
While the pure ViT \cite{dosovitskiy2020image} can directly achieve 89.9\% accuracy, our TransFG constantly gets 0.9\% performance gain compared to ViT and reaches 90.8\% accuracy which outperforms SOTA by \textbf{1.6\%}.

\begin{table}[]
    \small
    \centering
    \caption{Comparison of different methods on iNat2017.}
    \label{tab:inat}
    \begin{tabular}{c|c|c}
    \hline
    Method & Backbone & iNat2017 \\ \hline
    ResNet152 & ResNet152 & 59.0 \\
    IncResNetV2 & IncResNetV2 & 67.3 \\ 
    TASN & ResNet101 & 68.2 \\ \hline
    ViT & ViT-B\_16 & 68.7 \\ 
    TransFG & ViT-B\_16 & \textbf{71.7} \\ \hline
    \end{tabular}
\end{table}

iNat2017 is a large-scale dataset for fine-grained species recognition. Most previous methods do not report results on iNat2017 because of the computational complexity of the multi-crop, multi-scale and multi-stage optimization. With the simplicity of our model pipeline, we are able to scale TransFG well to big datasets and evaluate the performance which is shown in Table \ref{tab:inat}. This dataset is very challenging for mining meaningful object parts and the background is very complicated as well. We find that Vision Transformer structure outperforms ResNet structure a lot in these large challenging datasets. ViT outperformes ResNet152 by nearly 10\% and similar phenomenon can also be observed in iNat2018 and iNat2019. Our TransFG is the only method to achieve above 70\% accuracy with input size of 304 and outperforms SOTA with a large margin of \textbf{3.5\%}.

For the just ended iNat2021 competition which contains 10,000 species, 2.7M training images, our TransFG achieves very high single model accuracy of 91.3\%. (The final performance was obtained by ensembling many different models along with multi-modality processing) As far as we know, at least two of the Top5 teams in the final leaderboard adopted TransFG as one of their ensemble models. This clear proves that our model can be further extended to large-scale challenging scenarios besides academy datasets.

\subsection{Ablation Study}
\label{sec:ablation}

We conduct ablation studies on our TransFG pipeline to analyze how its variants affect the fine-grained visual classification result. All ablation studies are done on CUB-200-2011 dataset while the same phenomenon can be observed on other datasets as well.

\begin{table}[]
    \small
    \centering
    \caption{Ablation study on split way of image patches on CUB-200-2011 dataset.}
    \label{tab:absplit}
    \begin{tabular}{|c|c|c|c|}
    \hline
    Method & Patch Split & Accuracy (\%) & Training Time (h)\\ \hline
    ViT & Non-Overlap & 90.3 & 1.30 \\ 
    ViT & Overlap & \textbf{90.5} & 3.38  \\ \hline
    TransFG & Non-Overlap & 91.5 & 1.98 \\
    TransFG & Overlap & \textbf{91.7} & 5.38 \\ \hline
    \end{tabular}
\end{table}

\noindent \textbf{Influence of image patch split method.} We investigate the influence of our overlapping patch split method through experiments with standard non-overlapping patch split. As shown in Table \ref{tab:absplit}, both on the pure Vision Transformer and our improved TransFG framework, the overlapping split method bring consistently improvement, i.e., 0.2\% for both frameworks. The additional computational cost introduced by this is also affordable as shown in the fourth column. 

\begin{table}[ht]
    \small
    \centering
    \caption{Ablation study on Part Selection Module (PSM) on CUB-200-2011 dataset.}
    \label{tab:abpsm}
    \begin{tabular}{|c|c|}
    \hline
    Method & Accuracy (\%) \\ \hline
    ViT & 90.3 \\ 
    TransFG & \textbf{91.0} \\ \hline
    \end{tabular}
\end{table}

\noindent \textbf{Influence of Part Selection Module.} As shown in Table \ref{tab:abpsm}, by applying the Part Selection Module (PSM) to select discriminative part tokens as the input for the last Transformer layer, the performance of the model improves from 90.3\% to 91.0\%. We argue that this is because in this way, we sample the most discriminative tokens as input which explicitly throws away some useless tokens and force the network to learn from the important parts.

\begin{table}[]
    \small
    \centering
    \caption{Ablation study on contrastive loss on CUB-200-2011 dataset.}
    \label{tab:abdup}
    \begin{tabular}{|c|c|c|c|}
    \hline
    Method & Contrastive Loss & Acc (\%) \\ \hline
    ViT & & 90.3 \\
    ViT & \checkmark & \textbf{90.7} \\ \hline
    TransFG & & 91.0 \\ 
    TransFG & \checkmark & \textbf{91.5} \\ \hline
    \end{tabular}
\end{table}

\begin{figure}[h]
    \centering
    \includegraphics[width=\linewidth,height=4cm]{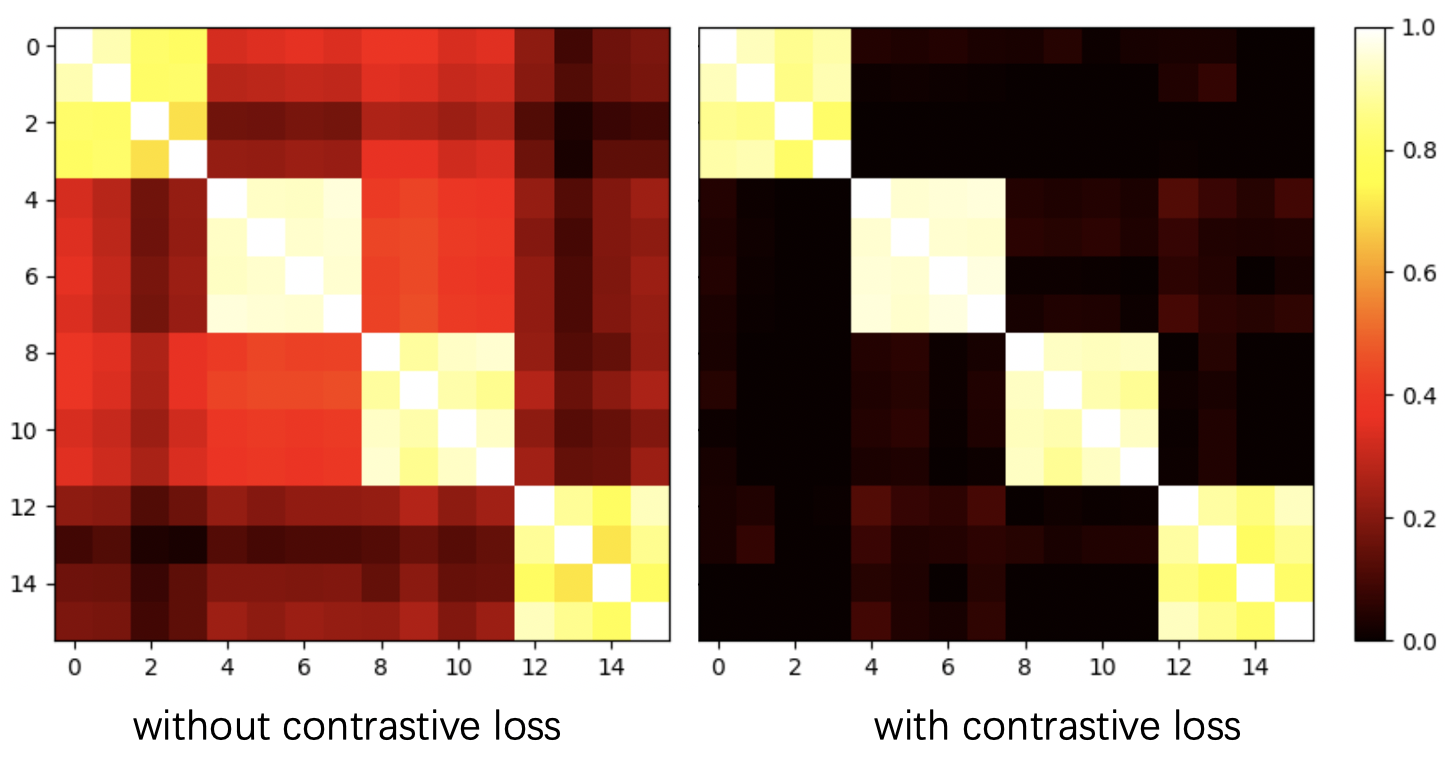}
    \caption{Illustration of contrastive loss. Confusion matrices without and with contrastive loss of a batch with four classes where each contains four samples are shown. The metric of confusion matrix is cosine similarity. Best viewed in color.}
    \label{fig:confusion}
\end{figure}

\noindent \textbf{Influence of contrastive loss.} The comparisons of the performance with and without contrastive loss for both ViT and TransFG frameworks are shown in Table \ref{tab:abdup} to verify the effectiveness of it. We observe that with contrastive loss, the model obtains a big performance gain. Quantitatively, it increases the accuracy from 90.3\% to 90.7\% for ViT and 91.0\% to 91.5\% for TransFG. We argue that this is because contrastive loss can effectively enlarge the distance of representations between similar sub-categories and decrease that between the same categories which can be clearly seen in the comparison of confusion matrix in Fig \ref{fig:confusion}.

\begin{table}[]
    \small
    \centering
    \caption{Ablation study on value of margin $\alpha$ on CUB-200-2011 dataset.}
    \label{tab:abalpha}
    \begin{tabular}{|c|c|c|}
    \hline
    Method & Value of $\alpha$ & Accuracy (\%) \\ \hline
    TransFG & 0 & 91.1 \\ 
    TransFG & 0.2 & 91.4 \\ 
    TransFG & 0.4 & \textbf{91.7} \\ 
    TransFG & 0.6 & 91.5 \\ \hline
    \end{tabular}
\end{table}

\begin{figure*}[ht]
    \centering
    \begin{subfigure}{0.16\linewidth}
        \centering
		\includegraphics[width=\linewidth]{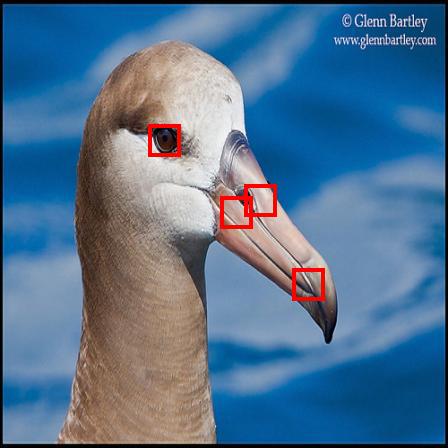}
    \end{subfigure}
    \begin{subfigure}{0.16\linewidth}
        \centering
		\includegraphics[width=\linewidth]{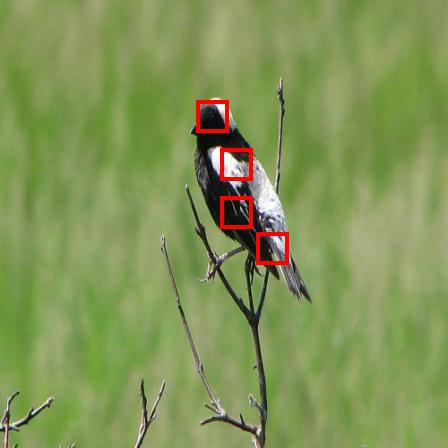}
    \end{subfigure}
    \begin{subfigure}{0.16\linewidth}
        \centering
		\includegraphics[width=\linewidth]{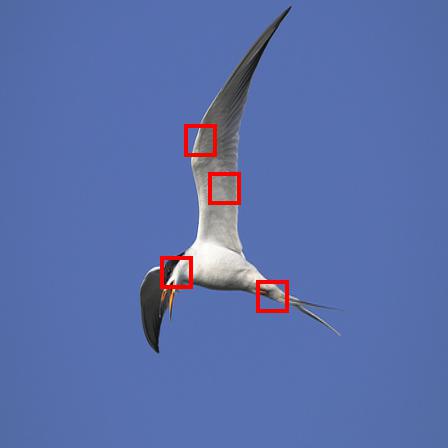}
    \end{subfigure}
    \begin{subfigure}{0.16\linewidth}
        \centering
		\includegraphics[width=\linewidth]{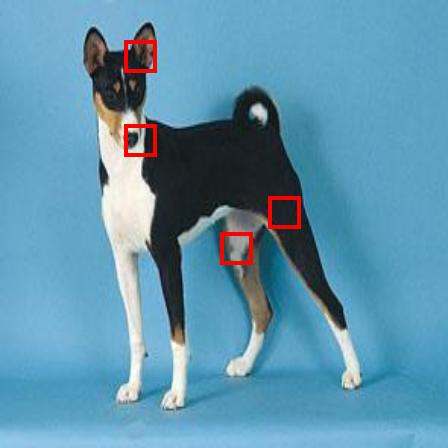}
    \end{subfigure}
    \begin{subfigure}{0.16\linewidth}
        \centering
		\includegraphics[width=\linewidth]{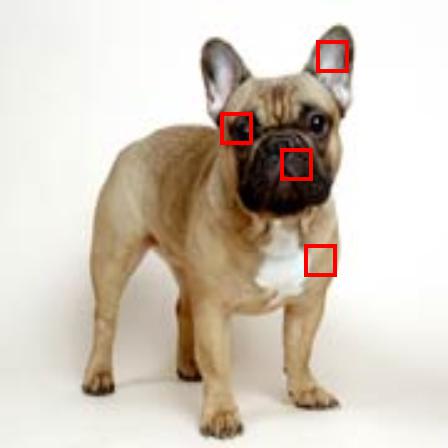}
    \end{subfigure}
    \begin{subfigure}{0.16\linewidth}
        \centering
		\includegraphics[width=\linewidth]{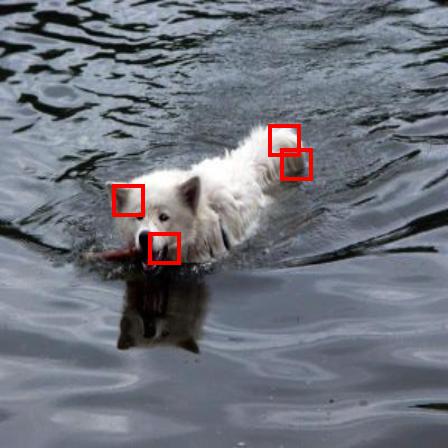}
    \end{subfigure}
    \begin{subfigure}{0.16\linewidth}
        \centering
		\includegraphics[width=\linewidth]{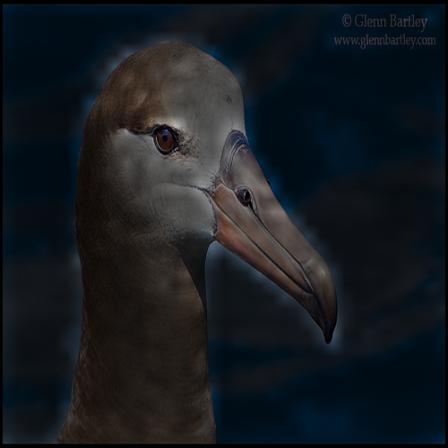}
    \end{subfigure}
    \begin{subfigure}{0.16\linewidth}
        \centering
		\includegraphics[width=\linewidth]{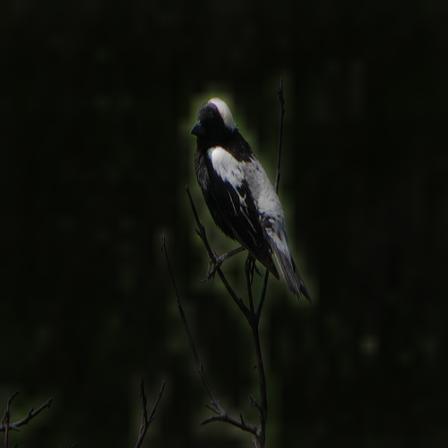}
    \end{subfigure}
    \begin{subfigure}{0.16\linewidth}
        \centering
		\includegraphics[width=\linewidth]{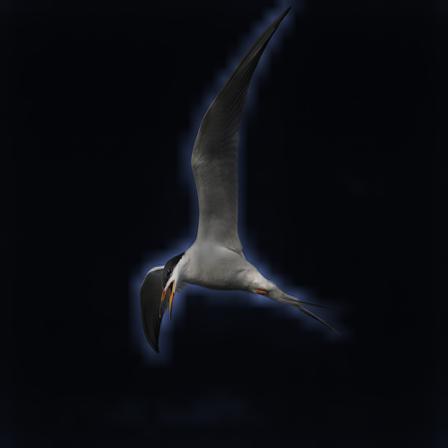}
    \end{subfigure}
    \begin{subfigure}{0.16\linewidth}
        \centering
		\includegraphics[width=\linewidth]{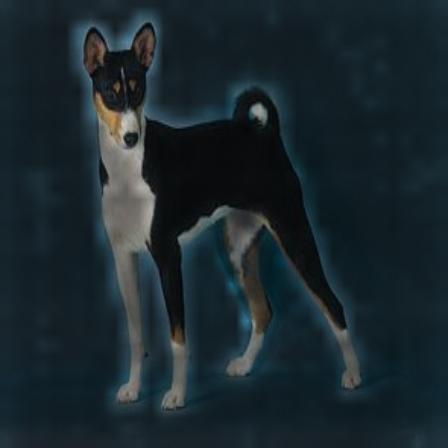}
    \end{subfigure}
    \begin{subfigure}{0.16\linewidth}
        \centering
		\includegraphics[width=\linewidth]{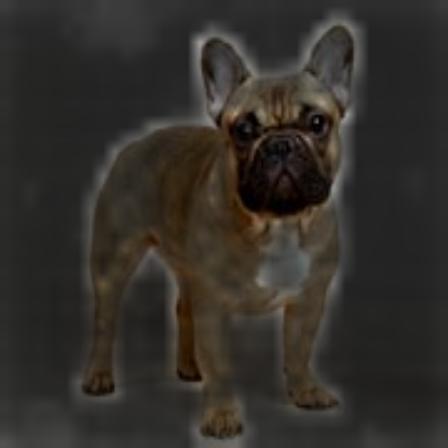}
    \end{subfigure}
    \begin{subfigure}{0.16\linewidth}
        \centering
		\includegraphics[width=\linewidth]{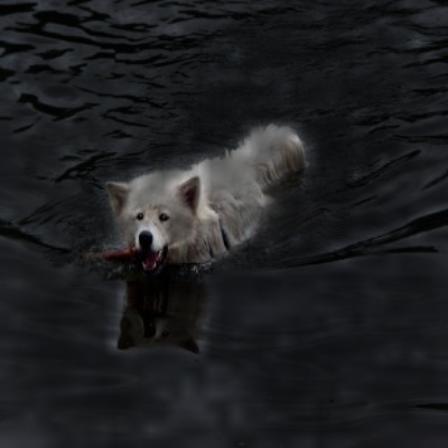}
    \end{subfigure}
    \begin{subfigure}{0.16\linewidth}
        \centering
		\includegraphics[width=\linewidth]{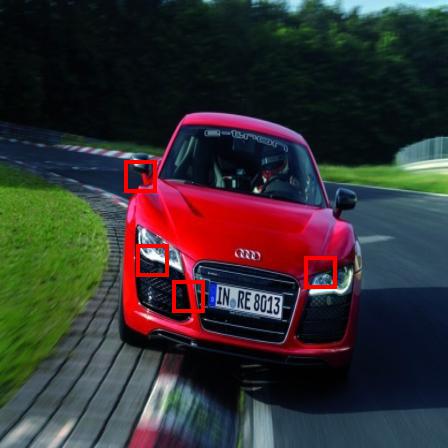}
    \end{subfigure}
    \begin{subfigure}{0.16\linewidth}
        \centering
		\includegraphics[width=\linewidth]{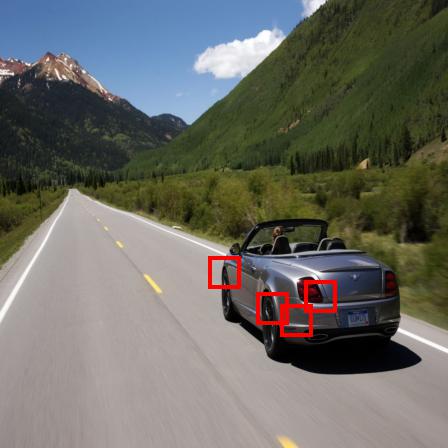}
    \end{subfigure}
    \begin{subfigure}{0.16\linewidth}
        \centering
		\includegraphics[width=\linewidth]{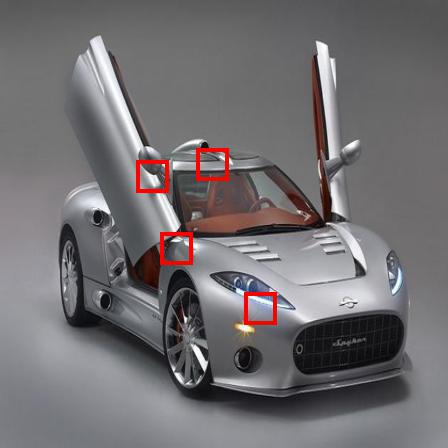}
    \end{subfigure}
    \begin{subfigure}{0.16\linewidth}
        \centering
		\includegraphics[width=\linewidth]{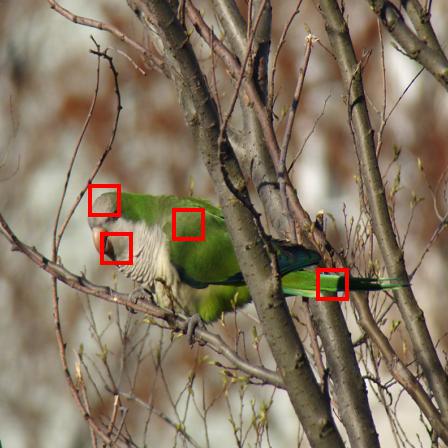}
    \end{subfigure}
    \begin{subfigure}{0.16\linewidth}
        \centering
		\includegraphics[width=\linewidth]{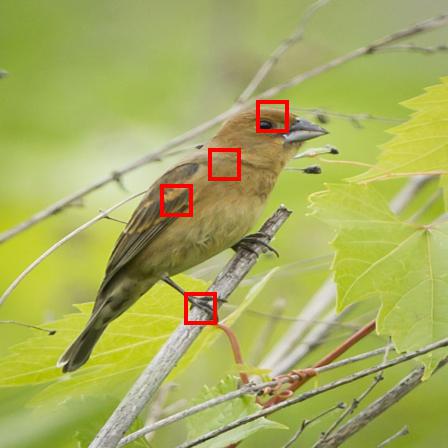}
    \end{subfigure}
    \begin{subfigure}{0.16\linewidth}
        \centering
		\includegraphics[width=\linewidth]{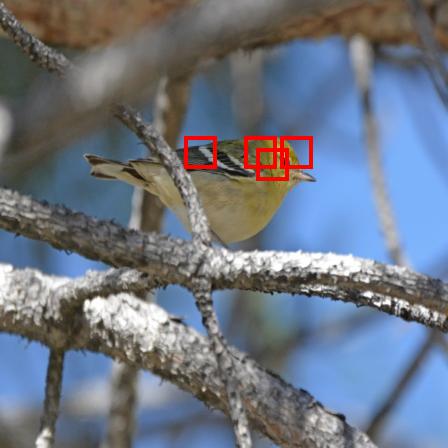}
    \end{subfigure}
    \begin{subfigure}{0.16\linewidth}
        \centering
		\includegraphics[width=\linewidth]{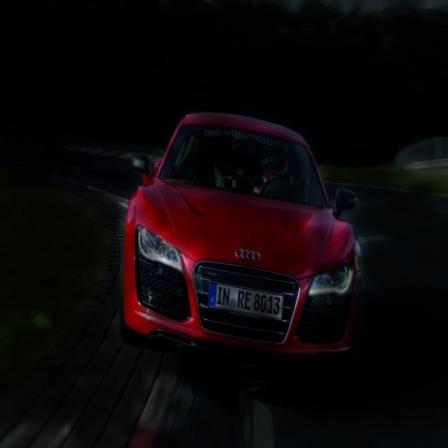}
    \end{subfigure}
    \begin{subfigure}{0.16\linewidth}
        \centering
		\includegraphics[width=\linewidth]{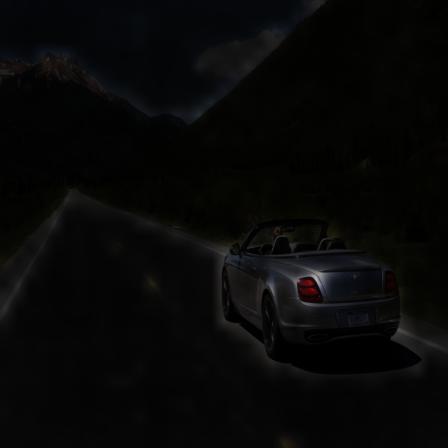}
    \end{subfigure}
    \begin{subfigure}{0.16\linewidth}
        \centering
		\includegraphics[width=\linewidth]{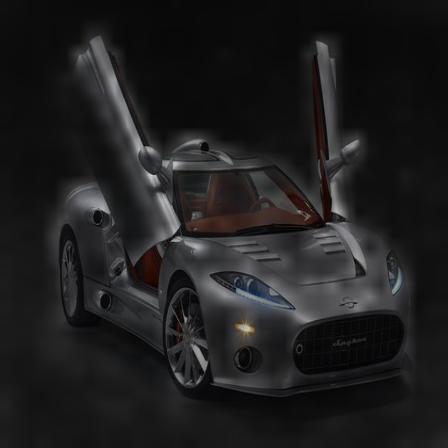}
    \end{subfigure}
    \begin{subfigure}{0.16\linewidth}
        \centering
		\includegraphics[width=\linewidth]{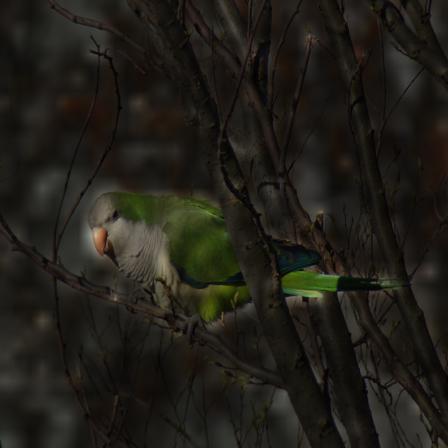}
    \end{subfigure}
    \begin{subfigure}{0.16\linewidth}
        \centering
		\includegraphics[width=\linewidth]{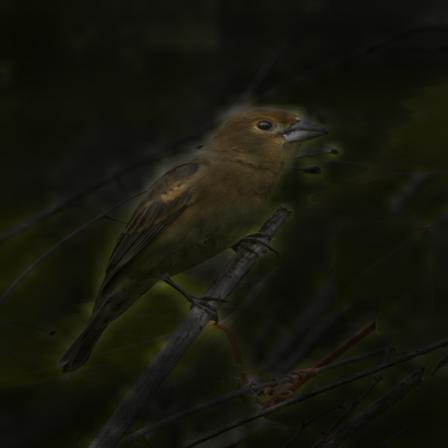}
    \end{subfigure}
    \begin{subfigure}{0.16\linewidth}
        \centering
		\includegraphics[width=\linewidth]{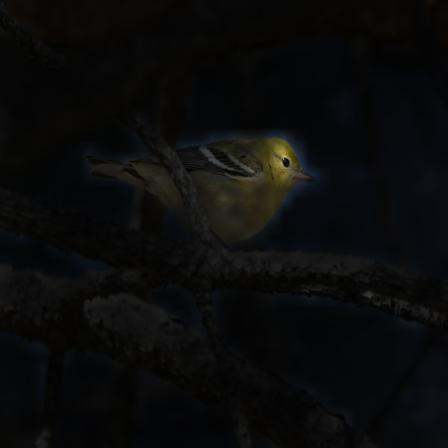}
    \end{subfigure}
    \caption{Visualization results of TransFG on CUB-200-2011, Stanford Dogs, Stanford Cars and NABirds datasets. Two kinds of visualization are given, where the first and the third row show the selected Top-4 token positions while the second and fourth rows show the overall global attention maps. See examples from NABirds dataset where birds are sitting on twigs. The bird parts are lighted while the occluded twigs are ignored. Best viewed in color.}
    \label{fig:vis}
\end{figure*}

\noindent \textbf{Influence of margin $\alpha$.} The results of different setting of the margin $\alpha$ in Eq \ref{equ:con} is shown in Table \ref{tab:abalpha}. We find that a small value of $\alpha$ will lead the training signals dominated by easy negatives thus decrease the performance while a high value of $\alpha$ hinder the model to learn sufficient information for increasing the distances of hard negatives. Empirically, we find 0.4 to be the best value of $\alpha$ in our experiments. 

\subsection{Qualitative Analysis}
\label{sec:qual}

We show the visualization results of proposed TransFG on the four benchmarks in Fig \ref{fig:vis}. We randomly sample three images from each dataset. Two kinds of visualizations are presented. The first and the third row of Fig \ref{fig:vis} illustrated the selected tokens positions. For better visualization results, we only draw the Top-4 image patches (ranked by the attention score) and enlarge the square of the patches by two times while keeping the center positions unchanged. The second and fourth rows show the overall attention map of the whole image where we use the same attention integration method as described above to first integrate the attention weights of all layers followed by averaging the weights of all heads to obtain a single attention map. The lighter a region is, the more important it is. From the figure, we can see that our TransFG successfully captures the most important regions for an object, i.e., head, wings, tail for birds; ears, eyes, legs for dogs; lights, doors for cars. At the same time, our overall attention map maps the entire object precisely even in complex backgrounds and it can even serves as a segmentation mask in some simple scenarios. These visualization results clearly prove the interpretability of our proposed method.
\section{Conclusion}

In this work, we propose a novel fine-grained recognition framework TransFG and achieve state-of-the-art results on four common fine-grained benchmarks. We exploit self-attention mechanism to capture the most discriminative regions. Compared to bounding boxes produced by other methods, our selected image patches are much smaller thus becoming more meaningful by showing what regions really contribute to the fine-grained classification. The effectiveness of such small image patches also comes from the Transformer Layer to handle the inner relationships between these regions instead of relying on each of them to produce results separately. Contrastive loss is introduced to increase the discriminative ability of the classification tokens. Experiments are conducted on both traditional academy datasets and large-scale competition datasets to prove the effectiveness of our model in multiple scenarios. Qualitative visualizations further show the interpretability of our method.

With the promising results achieved by TransFG, we believe that the transformer-based models have great potential on fine-grained tasks and our TransFG could be a starting point for future works.

{\small
\bibliography{egbib}
}

\end{document}